# Asymptotic Model Selection for Directed Networks with Hidden Variables*

**Dan Geiger**
Computer Science Department
Technion, Haifa 32000, Israel
dang@cs.technion.ac.il

**David Heckerman**
Microsoft Research, Bldg 9S
Redmond WA, 98052-6399
heckerma@microsoft.com

**Christopher Meek**
Carnegie-Mellon University
Department of Philosophy
meek@cmu.edu

## Abstract

We extend the Bayesian Information Criterion (BIC), an asymptotic approximation for the marginal likelihood, to Bayesian networks with hidden variables. This approximation can be used to select models given large samples of data. The standard BIC as well as our extension punishes the complexity of a model according to the dimension of its parameters. We argue that the dimension of a Bayesian network with hidden variables is the rank of the Jacobian matrix of the transformation between the parameters of the network and the parameters of the observable variables. We compute the dimensions of several networks including the naive Bayes model with a hidden root node.

## 1 Introduction

Learning Bayesian networks from data extends their applicability to situations where data is easily obtained and expert knowledge is expensive. Consequently, it has been the subject of much research in recent years (see e.g., Heckerman [1995] and Buntine [1996]). Researchers have pursued two types of approaches for learning Bayesian networks: one that uses independence tests to direct a search among valid models and another that uses a score to search for the best scored network—a procedure known as *model selection*. Scores based on exact Bayesian computations have been developed by (e.g.) Cooper and Herskovits (1992), Spiegelhalter et al. (1993), Buntine (1994), and Heckerman et al. (1995), and scores based on minimum description length (MDL) have been developed in Lam and Bacchus (1993) and Suzuki (1993).

We consider a Bayesian approach to model selection.

Suppose we have a set $\{X_1, \ldots, X_n\} = \mathbf{X}$ of discrete variables, and a set $\{\mathbf{x}_1, \ldots, \mathbf{x}_N\} = D$ of cases, where each case is an instance of some or of all the variables in $\mathbf{X}$. Let $(S, \boldsymbol{\theta}_s)$ be a Bayesian network, where $S$ is the network structure of the Bayesian network, a directed acyclic graph such that each node $X_i$ of $S$ is associated with a random variable $X_i$, and $\boldsymbol{\theta}_s$ is a set of parameters associated with the network structure. Let $S^h$ stand for the hypothesis that the true or objective joint distribution of $\mathbf{X}$ can be encoded in the network structure $S$. Then, a Bayesian measure of the goodness-of-fit of network structure $S$ to $D$ is $p(S^h|D) \propto p(S^h) p(D|S^h)$, where $p(D|S^h)$ is known as the *marginal likelihood of $D$ given $S^h$*.

The problem of model selection among Bayesian networks with hidden variables, that is, networks with variables whose values are not observed is more difficult than model selection among networks without hidden variables. First, the space of possible networks becomes infinite, and second, scoring each network is computationally harder because one must account for all possible values of the missing variables (Cooper and Herskovits, 1992). Our goal is to develop a Bayesian scoring approach for networks that include hidden variables. Obtaining such a score that is computationally effective and conceptually simple will allow us to select a model from among a set of competing models.

Our approach is to use an asymptotic approximation of the marginal likelihood. This asymptotic approximation is known as the Bayesian Information Criteria (BIC) (Schwarz, 1978), and is equivalent to Rissanen's (1987) minimum description length (MDL). Such an asymptotic approximation has been carried out for Bayesian networks by Herskovits (1991) and Bouckaert (1995) when no hidden variables are present. Bouckaert (1995) shows that the marginal likelihood of data $D$ given a network structure $S$ is given by

$$p(D|S^h) = H(S, D)N - 1/2 \dim(S) \log(N) + O(1) \quad (1)$$

where $N$ is the sample size of the data, $H(S,D)$ is the entropy of the probability distribution obtained by projecting the frequencies of observed cases into the conditional probability tables of the Bayesian network $S$, and $\dim(S)$ is the number of parameters in $S$. Eq. 1 reveals the qualitative preferences made by the Bayesian approach. First, with sufficient data, a network structure that is an I-map of the true distribution is more likely than a network structure that is not an I-map of the true distribution. Second, among all network structures that are I-maps of the true distribution, the one with the minimum number of parameters is more likely.

Eq. 1 was derived from an explicit formula for the probability of a network given data by letting the sample size $N$ run to infinity and using a Dirichlet prior for its parameters. Nonetheless, Eq. 1 does not depend on the selected prior. In Section 3, we use Laplace's method to rederive Eq. 1 without assuming a Dirichlet prior. Our derivation is a standard application of asymptotic Bayesian analysis. This derivation is useful for gaining intuition for the hidden-variable case.

In section 4, we provide an approximation to the marginal likelihood for Bayesian networks with hidden variables, and give a heuristic argument for this approximation using Laplace's method. We obtain the following equation:

$$\log p(S|D) \approx \log p(S|D, \hat{\boldsymbol{\theta}}_s) - 1/2 \dim(S, \hat{\boldsymbol{\theta}}_s) \log(N) \quad (2)$$

where $\hat{\boldsymbol{\theta}}_s$ is the *maximum likelihood* (ML) value for the parameters of the network and $\dim(S, \hat{\boldsymbol{\theta}}_s)$ is the dimension of $S$ at the ML value for $\boldsymbol{\theta}_s$. The dimension of a model can be interpreted in two equivalent ways. First, it is the number of free parameters needed to represent the parameter space near the maximum likelihood value. Second, it is the rank of the Jacobian matrix of the transformation between the parameters of the network and the parameters of the observable (non-hidden) variables. In any case, the dimension depends on the value of $\hat{\boldsymbol{\theta}}_s$, in contrast to Eq. 1, where the dimension is fixed throughout the parameter space.

In Section 5, we compute the dimensions of several network structures, including the naive Bayes model with a hidden class node. In Section 6, we demonstrate that the scoring function used in AutoClass sometimes diverges from $p(S|D)$ asymptotically. In Sections 7 and 8, we describe how our approach can be extended to Gaussian and sigmoid networks.

## 2 Background

We introduce the following notation for a Bayesian network. Let $r_i$ be the number of states of variable $X_i$, $\mathbf{Pa}_i$ be the set of variables corresponding to the parents of node $X_i$, and $q_i = \prod_{X_l \in \mathbf{Pa}_i} r_l$ be the number of states of $\mathbf{Pa}_i$. We use the integer $j$ to index the states of $\mathbf{Pa}_i$. That is, we write $\mathbf{Pa}_i = \mathbf{pa}_i^j$ to denote that the parents of $X_i$ are assigned its $j$th state. We use $\theta_{ijk}$ to denote the true probability or *parameter* that $X_i = x_i^k$ given that $\mathbf{Pa}_i = \mathbf{pa}_i^j$. Note that $\sum_{k=1}^{r_i} \theta_{ijk} = 1$. Also, we assume $\theta_{ijk} > 0$. In addition, we use $\boldsymbol{\theta}_{ij} = \{\theta_{ijk} | 1 \leq k \leq r_i\}$ to denote the parameters associated with node $i$ for a given instance of the parents $\mathbf{Pa}_i$, and $\boldsymbol{\theta}_i = \{\boldsymbol{\theta}_{ij} | 1 \leq j \leq q_i\}$ to denote the parameters associated with node $i$. Thus, $\boldsymbol{\theta}_s = \{\boldsymbol{\theta}_i | 1 \leq i \leq n\}$. When $S$ is unambiguous, we use $\boldsymbol{\theta}$ instead of $\boldsymbol{\theta}_s$.

To compute $p(D|S^h)$ in closed form, several assumptions are usually made. First, the data $D$ is assumed to be a random sample from some Bayesian network $(S, \boldsymbol{\theta}_s)$. Second, for each network structure, the parameter sets $\boldsymbol{\theta}_1, \ldots, \boldsymbol{\theta}_n$ are mutually independent (global independence [Spiegelhalter and Lauritzen, 1990]), and the parameter sets $\boldsymbol{\theta}_{i1}, \ldots, \boldsymbol{\theta}_{iq_i}$ for each $i$ are assumed to be mutually independent (local independence [Spiegelhalter and Lauritzen, 1990]). Third, if a node has the same parents in two distinct networks, then the distribution of the parameters associated with this node are identical in both networks (parameter modularity [Heckerman et al., 1994]). Fourth, each case is complete. Fifth, the prior distribution of the parameters associated with each node is Dirichlet—that is, $p(\boldsymbol{\theta}_{ij}|S^h) \propto \prod_k \theta_{ijk}^{\alpha_{ijk}}$ where $\alpha_{ijk}$ can be interpreted as the equivalent number of cases seen in which $X_i = x_i^k$ and $\mathbf{Pa}_i = \mathbf{pa}_i^j$.

Using these assumptions, Cooper and Herskovits (1992) obtained the following exact formula for the marginal likelihood:

$$p(D|S^h) = \prod_{i=1}^{n} \prod_{j=1}^{q_i} \frac{\Gamma(\alpha_{ij})}{\Gamma(\alpha_{ij} + N_{ij})} \prod_{k=1}^{r_i} \frac{\Gamma(\alpha_{ijk} + N_{ijk})}{\Gamma(\alpha_{ijk})}$$

where $N_{ijk}$ is the number of cases in $D$ in which $X_i = x_i^k$ and $\mathbf{Pa}_i = \mathbf{pa}_i^j$. We call this expression the *Cooper–Herskovits scoring function*.

The last two assumptions are made for the sake of convenience. Namely, the parameter distributions before and after data are seen are in the same family: the Dirichlet family. Geiger and Heckerman (1995) provide a characterization of the Dirichlet distribution, which shows that the fifth assumption is implied from the first three assumptions and from one additional assumption that if $S_1$ and $S_2$ are equivalent Bayesian networks (i.e., they represent the same sets of joint distributions), then the events $S_1^h$ and $S_2^h$ are equivalent as well (hypothesis equivalence [Heckerman et al., 1995]). This assumption was made explicit, because it does not hold for causal networks where two arcs with opposing directions correspond to distinct hypotheses [Heckerman, 1995a]. To satisfy these assumptions, Heckerman et al. (1995) show that one must use

$$\alpha_{ijk} = \alpha \, q(X_i = x_i^k, \mathbf{Pa}_i = \mathbf{pa}_i^j)$$

in the Cooper–Herskovits scoring function, where $q(X_1, \ldots, X_n)$ is the joint probability distribution of $\mathbf{X}$ obtained from an initial or prior Bayesian network specified by the user, and $\alpha$ is the user's effective sample size or confidence in the prior network.

The Cooper–Herskovits scoring function does not lend itself to a qualitative analysis. Nonetheless, by letting $N$ grow to infinity yet keeping $N_{ij}/N$ and $N_{ijk}/N$ finite, Eq. 1 can be derived by expanding $\Gamma(\cdot)$ using Sterling's approximation. This derivation hinges on the assumptions of global and local independence and on a Dirichlet prior, although, as we show, the result still holds without these assumptions. Intuitively, with a large sample size $N$, the data washes away any contribution of the prior.

## 3 Assymptotics Without Hidden Variables

We shall now rederive Herskovits' (1991) and Bouckaert's (1995) asymptotic result. The technique we use is Laplace's method, which is to expand the log likelihood of the data around the maximum likelihood value, and then approximate the peak using a multivariate-normal distribution.

Our derivation bypasses the need to compute $p(D_N|S^h)$ for data $D_N$ of a sample size $N$, which requires the assumptions discussed in the previous section. Instead, we compute $\lim_{N\to\infty} p(D_N|S^h)$. Furthermore, our derivation only assumes that the prior for $\boldsymbol{\theta}$ around the maximum likelihood value is positive. Finally, we argue in the next section that our derivation can be extended to Bayesian networks with hidden variables.

We begin by defining $f(\boldsymbol{\theta}) \equiv \log p(D_N|\boldsymbol{\theta}, S^h)$. Thus,

$$p(D_N|S^h) = \int p(D_N|\boldsymbol{\theta}, S^h) \, p(\boldsymbol{\theta}|S^h) \, d\boldsymbol{\theta} = $$
$$\int \exp\{f(\boldsymbol{\theta})\} \, p(\boldsymbol{\theta}|S^h) \, d\boldsymbol{\theta} \quad (3)$$

Assuming $f(\boldsymbol{\theta})$ has a maximum—the ML value $\hat{\boldsymbol{\theta}}$—we have $f'(\hat{\boldsymbol{\theta}}) = 0$. Using a Taylor-series expansion of $f(\boldsymbol{\theta})$ around the ML value, we get

$$f(\boldsymbol{\theta}) \approx f(\hat{\boldsymbol{\theta}}) + 1/2(\boldsymbol{\theta} - \hat{\boldsymbol{\theta}})f''(\hat{\boldsymbol{\theta}})(\boldsymbol{\theta} - \hat{\boldsymbol{\theta}}) \quad (4)$$

where $f''(\boldsymbol{\theta})$ is the Hessian of $f$—the square matrix of second derivatives with respect to every pair of variables $\{\theta_{ijk}, \theta_{i'j'k'}\}$. Consequently, from Eqs. 3 and 4,

$$\log p(D|S^h) \approx f(\hat{\boldsymbol{\theta}})+ \quad (5)$$
$$\log \int \exp\{1/2(\boldsymbol{\theta} - \hat{\boldsymbol{\theta}})f''(\boldsymbol{\theta})(\boldsymbol{\theta} - \hat{\boldsymbol{\theta}})\}p(\boldsymbol{\theta}|S^h)d\boldsymbol{\theta}$$

We assume that $-f''(\boldsymbol{\theta})$ is positive-definite, and that, as $N$ grows to infinity, the peak in a neighborhood around the maximum becomes sharper. Consequently, if we ignore the prior, we get a normal distribution around the peak. Furthermore, if we assume that the prior $p(\boldsymbol{\theta}|S^h)$ is not zero around $\hat{\boldsymbol{\theta}}$, then as $N$ grows it can be assumed constant and so removed from the integral in Eq. 5. The remaining integral is approximated by the formula for multivariate-normal distributions:

$$\int \exp\{1/2(\boldsymbol{\theta} - \hat{\boldsymbol{\theta}})f''(\boldsymbol{\theta})(\boldsymbol{\theta} - \hat{\boldsymbol{\theta}})\}d\boldsymbol{\theta} \approx$$
$$\sqrt{2\pi} \det\left[-f''(\hat{\boldsymbol{\theta}})\right]^{d/2} \quad (6)$$

where $d$ is the number of parameters in $\boldsymbol{\theta}$, $d = \prod_{i=1}^n (r_i - 1)q_i$. As $N$ grows to infinity, the above approximation becomes more precise because the entire mass becomes concentrated around the peak. Plugging Eq. 6 into Eq. 5 and noting that $\det\left[-f''(\hat{\boldsymbol{\theta}})\right]$ is proportional to $N$ yields the BIC:

$$p(D_N|S^h) \approx p(D_N|\hat{\boldsymbol{\theta}}, S^h) - d/2 \log(N) \quad (7)$$

A careful derivation in this spirit shows that the error in this approximation does not depend on $N$ [Schwarz, 1978].

For Bayesian networks, the function $f(\boldsymbol{\theta})$ is known. Thus, all the assumptions about this function can be verified. First, we note that $f''(\boldsymbol{\theta})$ is a block diagonal matrix where each block $A_{ij}$ corresponds to variable $X_i$ and a particular instance $j$ of $\mathbf{Pa}_i$, and is of size $(r_i - 1)^2$. Let us examine one such $A_{ij}$. To simplify notation, assume that $X_i$ has three states. Let $w_1, w_2$ and $w_3$ denote $\theta_{ijk}$ for $k = 1, 2, 3$, where $i$ and $j$ are fixed. We consider only those cases in $D_N$ where $\mathbf{Pa}_i = j$, and examine only the observations of $X_i$. Let $D'_N$ denote the set of $N$ values of $X_i$ obtained in this process. With each observation, we associate two indicator functions $x_i$ and $y_i$. The function $x_i$ is one if $X_i$ gets its first value in case $i$ and is zero otherwise. Similarly, $y_i$ is one if $X_i$ gets its second value in case $i$ and is zero otherwise.

The log likelihood function of $D'_N$ is given by

$$\lambda(w_1, w_2) = \log \prod_{i=1}^{N} w_1^{x_i} w_2^{y_i} (1 - w_1 - w_2)^{1-x_i-y_i} \quad (8)$$

To find the maximum, we set the first derivative of this function to zero. The resulting equations are called the maximum likelihood equations:

$$\lambda_{w_1}(w_1, w_2) = \sum_{i=1}^{N} \left[ \frac{x_i}{w_1} - \frac{1 - x_i - y_i}{1 - w_1 - w_2} \right] = 0$$

$$\lambda_{w_2}(w_1, w_2) = \sum_{i=1}^{N} \left[ \frac{y_i}{w_2} - \frac{1 - x_i - y_i}{1 - w_1 - w_2} \right] = 0$$

The only solution to these equations is given by $w_1 = \overline{x} = \sum_i x_i/N$, $w_2 = \overline{y} = \sum_i y_i/N$, which is the maximum likelihood value. The Hessian of $\lambda(w_1, w_2)$ at the ML value is given by

$$\lambda''(w_1, w_2) = \begin{pmatrix} \lambda''_{w_1 w_1} & \lambda''_{w_1 w_2} \\ \lambda''_{w_1 w_1} & \lambda''_{w_2 w_2} \end{pmatrix} =$$

$$-N \begin{pmatrix} \frac{1}{\overline{x}} + \frac{1}{1-\overline{x}-\overline{y}} & \frac{1}{1-\overline{x}-\overline{y}} \\ \frac{1}{1-\overline{x}-\overline{y}} & \frac{1}{\overline{y}} + \frac{1}{1-\overline{x}-\overline{y}} \end{pmatrix} \quad (9)$$

This Hessian matrix decomposes into the sum of two matrices. One matrix is a diagonal matrix with positive numbers $1/\overline{x}$ and $1/\overline{y}$ on the diagonal. The second matrix is a constant matrix in which all elements equal the positive number $1/(1 - \overline{x} - \overline{y})$. Because these two matrices are positive and non-negative definite, respectively, the Hessian is positive definite. This argument also holds when $X_i$ has more than three values.

Because the maximum likelihood equation has a single solution, and the Hessian is positive definite, and because as $N$ increases the peak becomes sharper (Eq.9), all the conditions for the general derivation of the BIC are met. Plugging the maximum likelihood value into Eq. 7, which is correct to $O(1)$, yields Eq. 1.

## 4 Assymptotics With Hidden Variables

Let us now consider the situation where $S$ contains hidden variables. In this case, we can not use the derivation in the previous section, because the log-likelihood function $\log p(D_N|S^h, \boldsymbol{\theta})$ does not necessarily tend toward a *peak* as the sample size increases. Instead, the log-likelihood function can tend toward a *ridge*. Consider, for example, a network with one arc $H \to X$ where $H$ has two values $h$ and $\overline{h}$ and $X$ has two values $x$ and $\overline{x}$. Assume that only values of $X$ are observed—that is, $H$ is hidden. Then, the likelihood function is given by $\prod_i w^{x_i}(1-w)^{1-x_i}$ where

$w = \theta_h \theta_{x|h} + (1-\theta_h)\theta_{x|\overline{h}}$, and $x_i$ is the indicator function that equals one if $X$ gets value $x$ in case $i$ and zero otherwise. The parameter $w$ is the true probability that $X = x$ unconditionally. The ML value is unique in terms of $w$: it attains its maximum when $w = \sum_i x_i/N$. Nonetheless, any solution for $\boldsymbol{\theta}$ to the equation

$$\sum_i x_i/N = \theta_h \theta_{x|h} + (1-\theta_h)\theta_{x|\overline{h}}$$

will maximize the likelihood of the data. In this sense, the network structure $H \to X$ has only one non-redundant parameter. In this section, we provide an informal argument describing how to identify a set of non-redundant parameters for any Bayesian network with hidden variables.

Given a Bayesian network for domain $\mathbf{X}$ with observable variables $\mathbf{O} \subset \mathbf{X}$, let $W = \{w_{\mathbf{o}}|\mathbf{o} \in \mathbf{O}\}$ denote the parameters of the true joint probability distribution of $\mathbf{O}$. Corresponding to every value of $\boldsymbol{\theta}$ is a value of $W$. That is, $S$ defines a (smooth) map $g$ from $\boldsymbol{\theta}$ to $W$. The range of $g$ is a curved manifold $M$ in the space defined by $W$.[1] Now, consider $g(\hat{\boldsymbol{\theta}})$, the image of all ML values of $\boldsymbol{\theta}$. In a small region around $g(\hat{\boldsymbol{\theta}})$, the manifold $M$ will resemble Euclidean space with some dimension $d$. That is, in a small region around $g(\hat{\boldsymbol{\theta}})$, $M$ will look like $R^d$ with orthogonal coordinates $\Phi = \{\phi_1, \ldots, \phi_d\}$. Thus, the log-likelihood function written as a function of $\Phi$—$\log p(D_N|\Phi)$—will become peaked as the sample size increases, and we can apply the BIC approximation:

$$\log p(D_N|S^h) \approx \log p(D_N|\hat{\Phi}, S^h) - \frac{d}{2} \log N \quad (10)$$

Note that $\log p(D_N|\hat{\Phi}, S^h) = \log p(D_N|\hat{\boldsymbol{\theta}}, S^h)$.

It remains to understand what $d$ is and how it can be found. When considering a linear transformation $j : R^n \to R^m$, the transformation is a matrix of size $n \times m$. The dimension $d$ of the image of $j$ equals the rank of the matrix. When $k : R^n \to R^m$ is a smooth mapping, it can be approximated locally as a linear transformation, where the Jacobian matrix $J(\mathbf{x})$ serves as the linear transformation matrix for the neighborhood of $\mathbf{x} \in R^n$. The dimension of the image of $k$ in a small region around $k(\mathbf{x})$ is the rank of $J(\mathbf{x})$ (Spivak, 1979). This observation holds when the rank of the Jacobian matrix does not change in a small ball around $\mathbf{x}$, in which case $\mathbf{x}$ is called a *regular point*.

Returning to our problem, the mapping from $\boldsymbol{\theta}$ to $W$ is a polynomial function of $\boldsymbol{\theta}$. Thus, as the next theorem shows, the rank of the Jacobian matrix $\left[ \frac{\partial \boldsymbol{\theta}}{\partial W} \right]$ is almost everywhere some fixed constant $d$, which we call the *regular rank* of the Jacobian matrix. This rank is the number of non-redundant parameters of $S$—that is, the dimension of $S$.

---

[1] For terminology and basic facts in differential geometry, see Spivak (1979).

**Theorem 1** *Let $\boldsymbol{\theta}$ be the parameters of a network $S$ for variables $\mathbf{X}$ with observable variables $\mathbf{O} \subset \mathbf{X}$. Let $W$ be the parameters of the true joint distribution of the observable variables. If each parameter in $W$ is a polynomial function of $\boldsymbol{\theta}$, then* $\operatorname{rank}\left[\frac{\partial \boldsymbol{\theta}}{\partial W}(\boldsymbol{\theta})\right] = d$ *almost everywhere, where $d$ is a constant.*

**Proof:** Because the mapping from $\boldsymbol{\theta}$ to $W$ is polynomial, each entry in the matrix $J(\boldsymbol{\theta}) = \left[\frac{\partial \boldsymbol{\theta}}{\partial W}(\boldsymbol{\theta})\right]$ is a polynomial in $\boldsymbol{\theta}$. When diagonalizing $J$, the leading elements of the first $d$ lines remain polynomials in $\boldsymbol{\theta}$, whereas all other lines, which are dependent given every value of $\boldsymbol{\theta}$, become identically zero. The rank of $J(\boldsymbol{\theta})$ falls below $d$ only for values of $\boldsymbol{\theta}$ that are roots of some of the polynomials in the diagonalized matrix. The set of all such roots has measure zero. □

Our heuristic argument for Eq. 10 does not provide us with the error term. If the image manifold is too curved, it might be possible that the local region will never become "sufficiently flat" to obtain an $O(1)$ bound on the error of the approximate marginal likelihood. We conjecture that, for manifolds corresponding to Bayesian networks with hidden variables, the local region will always be sufficiently flat. Researchers have shown that $O(1)$ bounds are attainable for a variety of statistical models (e.g., Schwarz, 1978, and Haughton, 1988). Although the arguments of these researchers do not directly apply to our case, it may be possible to extend their methods to prove our conjecture.

## 5  Computations of the Rank

We have argued that the second term of the BIC for Bayesian networks with hidden variables is the rank of the Jacobian matrix of the transformation between the parameters of the network and the parameters of the observable variables. In this section, we explain how to compute this rank, and demonstrate the approach with several examples.

Theorem 1 suggests a random algorithm for calculating the rank. Compute the Jacobian matrix $J(\boldsymbol{\theta})$ symbolically from the equation $W = g(\boldsymbol{\theta})$. This computation is possible since $g$ is a vector of polynomials in $\boldsymbol{\theta}$. Then, assign a random value to $\boldsymbol{\theta}$ and diagonalize the numeric matrix $J(\boldsymbol{\theta})$. Theorem 1 guarantees that, with probability 1, the resulting rank is the regular rank of $J$. For every network, select—say—ten values for $\boldsymbol{\theta}$, and determine $r$ to be the maximum of the resulting ranks. In all our experiments, *none* of the randomly chosen values for $\boldsymbol{\theta}$ accidentally reduced the rank.

We now demonstrate the computation of the needed rank for a naive Bayes model with one hidden variable $H$ and two feature variables $X_1$ and $X_2$. Assume all three variables are binary. The set of parameters $W = g(\boldsymbol{\theta})$ is given by

$$w_{x_1 x_2} = \theta_h \theta_{x_1|h} \theta_{x_2|h} + (1-\theta_h)\theta_{x_1|\bar{h}}\theta_{x_2|\bar{h}}$$
$$w_{\bar{x}_1 x_2} = \theta_h (1-\theta_{x_1|h}) \theta_{x_2|h} + (1-\theta_h)(1-\theta_{x_1|\bar{h}})\theta_{x_2|\bar{h}}$$
$$w_{x_1 \bar{x}_2} = \theta_h \theta_{x_1|h} (1-\theta_{x_2|h}) + (1-\theta_h)\theta_{x_1|\bar{h}}(1-\theta_{x_2|\bar{h}})$$

The $3 \times 5$ Jacobian matrix for this transformation is given in Figure 5 where $\theta_{\bar{x}_i|h} = 1 - \theta_{x_i|h}$ ($i = 1, 2$). The columns correspond to differentiation with respect to $\theta_{x_1|h}, \theta_{x_2|h}, \theta_{x_1|\bar{h}}, \theta_{x_2|\bar{h}}$ and $\theta_h$, respectively. A symbolic computation of the rank of this matrix can be carried out; and it shows that the regular rank is equal to the dimension of the matrix—namely, 3. Nonetheless, as we have argued, in order to compute the regular rank, one can simply choose random values for $\boldsymbol{\theta}$ and diagonalize the resulting numerical matrix. We have done so for naive Bayes models with one binary hidden root node and $n \le 7$ binary observable non-root nodes. The size of the associated matrices is $(1+2n) \times (2^n - 1)$. The regular rank for $n = 3, \ldots, 7$ was found to be $1 + 2n$. We conjecture that $1 + 2n$ is the regular rank for all $n > 2$. For $n = 1, 2$, the rank is 1 and 3, respectively, which is the size of the full parameter space over one and two binary variables. The rank can not be greater than $1 + 2n$ because this is the maximum possible dimension of the Jacobian matrix. In fact, we have proven a lower bound of $2n$ as well.

**Theorem 2** *Let $S$ be a naive Bayes model with one binary hidden root node and $n > 2$ binary observable non-root nodes. Then*

$$2n \le r \le 2n + 1$$

*where $r$ is the regular rank of the Jacobian matrix between the parameters of the network and the parameters of the feature variables.*

The proof is obtained by diagonalizing the Jacobian matrix symbolically, and showing that there are at least $2n$ independent lines.

The computation for $3 \le n \le 7$ shows that, for naive Bayes models with a binary hidden root node, there are no redundant parameters. Therefore, the best way to represent a probability distribution that is representable by such a model is to use the network representation explicitly.

Nonetheless, this result does not hold for all models. For example, consider the following $W$ *structure*:

$$A \to C \leftarrow H \to D \leftarrow B$$

$$\begin{pmatrix} \theta_h\theta_{x_2|h} & \theta_h\theta_{x_1|h} & (1-\theta_h)\theta_{x_2|\bar{h}} & (1-\theta_h)\theta_{x_1|\bar{h}} & \theta_{x_1|h}\theta_{x_2|h} - \theta_{x_1|\bar{h}}\theta_{x_2|\bar{h}} \\ -\theta_h\theta_{x_2|h} & \theta_h\theta_{\bar{x}_1|h} & -(1-\theta_h)\theta_{x_2|\bar{h}} & (1-\theta_h)\theta_{\bar{x}_1|\bar{h}} & \theta_{\bar{x}_1|h}\theta_{x_2|h} - \theta_{\bar{x}_1|\bar{h}})\theta_{x_2|\bar{h}} \\ (1-\theta_h\theta_{x_2|h}) & -\theta_h\theta_{x_1|h} & (1-\theta_h)\theta_{\bar{x}_2|\bar{h}} & -(1-\theta_h)\theta_{x_1|\bar{h}} & \theta_{x_1|h}\theta_{\bar{x}_2|h} - \theta_{x_1|\bar{h}}\theta_{\bar{x}_2|\bar{h}} \end{pmatrix}$$

Figure 1: The Jacobian matrix for a naive Bayesian network with two binary feature nodes

where $H$ is hidden. Assuming all five variables are binary, the space over the observables is representable by 15 parameters, and the number of parameters of the network is 11. In this example, we could not compute the rank symbolically. Instead, we used the following Mathematica code.

There are 16 functions (only 15 are independent) defined by $W = g(\boldsymbol{\theta})$. In the Mathematica code, we use $fijkl$ for the true joint probability $w_{a=i,b=j,c=k,d=l}$, $cij$ for the true conditional probability $\theta_{c=0|a=i,h=j}$, $dij$ for $\theta_{d=0|b=i,h=j}$, $a$ for $\theta_{a=0}$, $b$ for $\theta_{b=0}$, and $h0$ for $\theta_{h=0}$.

The first function is given by

$f0000\,[a\_, b\_, h0\_, c00\_, \ldots, c11\_, d00\_, \ldots, d11\_] :=$
$\quad a * b * (h0 * c00 * d00 + (1 - h0) * c01 * d01)$

and the other functions are similarly written. The Jacobian matrix is computed by the command *Outer*, which has three arguments. The first is $D$ which stands for the differentiation operator, the second is a set of functions, and the third is a set of variables.

$J\,[a\_, b\_, h0\_, c00\_, \ldots, c11\_, d00\_, \ldots, d11\_] :=$
$\quad Outer[D, \{f0000\,[a, b, h0, c00, c01, \ldots, d11],$
$\quad f0001\,[a, b, h0, c00, \ldots, c11, d00, \ldots, d11],$
$\quad \ldots,$
$\quad f1111\,[a, b, h0, c00, \ldots, c11, d00, \ldots, d11]\},$
$\quad \{a, b, h0, c00, c01, c10, c11, d00, d01, d10, d11\}]$

The next command produces a diagonalized matrix at a random point with a precision of 30 decimal digits. This precision was selected so that matrix elements equal to zero would be correctly identified as such.

$N[RowReduce[J[a, b, h0, c00, \ldots, c11, d00, \ldots, d11]/.\{$
$\quad a \to \text{Random}[\text{Integer}, \{1, 999\}]/1000,$
$\quad b \to \text{Random}[\text{Integer}, \{1, 999\}]/1000,$
$\quad \ldots,$
$\quad d11 \to \text{Random}[\text{Integer}, \{1, 999\}]/1000\}], 30]$

The result of this Mathematica program was a diagonalized matrix with 9 non-zero rows and 7 rows containing all zeros. The same counts were obtained in ten runs of the program. Hence, the regular rank of this Jacobian matrix is 9 with probability 1.

The interpretation of this result is that, around almost every value of $\boldsymbol{\theta}$, one can locally represent the hidden W structure with only 9 parameters. In contrast, if we encode the distribution using the network parameters ($\boldsymbol{\theta}$) of the W structure, then we must use 11 parameters. Thus, two of the network parameters are locally redundant. The BIC approximation punishes this W structure according to its most efficient representation, which uses 9 parameters, and not according to the representation given by the W structure, which requires 11 parameters.

It is interesting to note that the dimension of the W structure is 10 if $H$ has three *or* four states, and 11 if $H$ has 5 states. We do not know how to predict when the dimension changes as a result of increasing the number of hidden states without computing the dimension explicitly. Nonetheless, the dimension can not increase beyond 12, because we can average out the hidden variable in the W structure (e.g., using arc reversals) to obtain another network structure that has only 12 parameters.

## 6 AutoClass

The AutoClass clustering algorithm developed by Cheeseman and Stutz (1995) uses a naive Bayes model.[2] Each state of the hidden root node $H$ represents a cluster or class; and each observable node represents a measurable feature. The number of classes $k$ is unknown a priori. AutoClass computes an approximation of the marginal likelihood of a naive Bayes model given the data using increasing values of $k$. When this probability reaches a peak for a specific $k$, that $k$ is selected as the number of classes.

Cheeseman and Stutz (1995) use the following formula to approximate the marginal likelihood:

$\log p(D|S) \approx$
$\quad \log p(D_c|S) + \log p(D|S, \hat{\boldsymbol{\theta}}_s) - \log p(D_c|S, \hat{\boldsymbol{\theta}}_s)$

where $D_c$ is a database consistent with the expected sufficient statistics as computed by the EM algorithm. Although Cheeseman and Stutz suggested this approximation in the context of simple AutoClass models, it can be used to score any Bayesian network with discrete variables as well as other models [Chickering and Heckerman, 1996]. We call this approximation the *CS scoring function*.

---

[2]The algorithm can handle conditional dependencies among continuous variables.

Using the BIC approximation for $p(D_c|S)$, we obtain

$$\log p(D|S) \approx \log p(D|S, \hat{\boldsymbol{\theta}}_s) - d'/2 \log N$$

where $d'$ is the number of parameters of the network. (Given a naive Bayes model with $k$ classes and $n$ observable variables each with $b$ states, $d' = nk(b-1) + k - 1$.) Therefore, the CS scoring function will converge asymptotically to the BIC and hence to $p(D|S)$ whenever $d'$ is equal to the regular rank of $S$ ($d$). Given our conjecture in the previous section, we believe that the CS scoring function will converge to $p(D|S)$ when the number of classes is two. Nonetheless, $d'$ is not always equal to $d$. For example, when $b = 2, k = 3$ and $n = 4$, the number of parameters is 14, but the regular rank of the Jacobian matrix is 13. We computed this rank using Mathematica as described in the previous section. Consequently, the CS scoring function will not always converge to $p(D|S)$.

This example is the only one that we have found so far; and we believe that incorrect results are obtained only for rare combinations of $b, k$ and $n$. Nonetheless, a simple modification to the CS scoring function yields an approximation that will asymptotically converge to $p(D|S)$:

$$\log p(D|S) \approx \log p(D_c|S) + \log p(D|S, \hat{\boldsymbol{\theta}}_s) - \\ \log p(D_c|S, \hat{\boldsymbol{\theta}}_s) - d/2 \log N + d'/2 \log N$$

Chickering and Heckerman (1996) show that this scoring function is often a better approximation for $p(D|S)$ than is the BIC.

## 7 Gaussian Networks

In this section, we consider the case where each of the variables $\{X_1, \ldots, X_n\} = \mathbf{X}$ are continuous. As before, let $(S, \boldsymbol{\theta}_s)$ be a Bayesian network, where $S$ is the network structure of the Bayesian network, and $\boldsymbol{\theta}_s$ is a set of parameters associated with the network structure. A Gaussian network is one in which the joint likelihood is that of a multivariate Gaussian distribution that is a product of local likelihoods. Each local likelihood is the linear regression model

$$p(x_i|\mathbf{pa}_i, \boldsymbol{\theta}_i, S) = N(m_i + \Sigma_{X_j \in \mathbf{Pa}_i} b_{ji} x_j, v_i)$$

where $N(\mu, v)$ is a normal (Gaussian) distribution with mean $\mu$ and variance $v > 0$, $m_i$ is a conditional mean of $X_i$, $b_{ji}$ is a coefficient that represents the strength of the relationship between variable $X_j$ and $X_i$, $v_i$ is a variance,[3] and $\boldsymbol{\theta}_i$ is the set of parameters consisting of $m_i$, $v_i$, and the $b_{ji}$. The parameters $\boldsymbol{\theta}_s$ of a Gaussian network with structure $S$ is the set of all $\boldsymbol{\theta}_i$.

To apply the techniques developed in this paper, we also need to specify the parameters of the observable variables. Given that the joint distribution is multivariate-normal and that multivariate-normal distributions are closed under marginalization, we only need to specify a vector of means for the observed variables and a covariance matrix over the observed variables. In addition, we need to specify how to transform the parameters of the network to the observable parameters. The transformation of the means and the transformation to obtain the observable covariance matrix can be accomplished via the *trek-sum rule* (for a discussion, see Glymour et al. 1987).

Using the trek-sum rule, it is easy to show that the observable parameters are all sums of products of the network parameters. Given that the mapping from $\boldsymbol{\theta}_s$ to the observable parameters is $W$ is a polynomial function of $\boldsymbol{\theta}$, it follows from Thm. 1 that the rank of the Jacobian matrix $\left[\frac{\partial \boldsymbol{\theta}_s}{\partial W}\right]$ is almost everywhere some fixed constant $d$, which we again call the *regular rank* of the Jacobian matrix. This rank is the number of non-redundant parameters of $S$—that is, the dimension of $S$.

Let us consider two Gaussian models. We use Mathematica code similar to the code in Section 5 to compute their dimensions, because we can not perform the computation symbolically. As in the previous experiments, none of the randomly chosen values of $\boldsymbol{\theta}_s$ accidentally reduces the rank.

Our first example is the naive-Bayes model

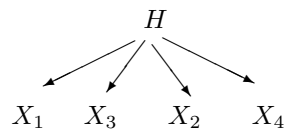

in which $H$ is the hidden variable and the $X_i$ are observed. There are 14 network parameters: 5 conditional variances, 5 conditional means, and 4 linear parameters. The marginal distribution for the observed variables also has 14 parameters: 4 means, 4 variances, and 6 covariances. Nonetheless, the analysis of the rank of the Jacobian matrix tells us that the dimension of this model is 12. This follows from the fact that this model imposes *tetrad* constraints (see Glymour et al. 1987). In this model the three tetrad constraints that

---

[3] $m_i$ is the mean of $X_i$ conditional on all parents being zero, $b_{ji}$ corresponds to the partial regression coefficient of $X_i$ on $X_j$ given the other parents of $X_i$, and $v_i$ corresponds to the residual variance of $X_i$ given the parents of $X_i$.

hold in the distribution over the observed variables are

$cov(X_1, X_2)cov(X_3, X_4) - cov(X_1, X_3)cov(X_2, X_4) = 0$
$cov(X_1, X_4)cov(X_2, X_3) - cov(X_1, X_3)cov(X_2, X_4) = 0$
$cov(X_1, X_4)cov(X_2, X_3) - cov(X_1, X_2)cov(X_3, X_4) = 0$

two of which are independent. These two independent tetrad constraints lead to the reduction of dimensionality.

Our second example is the $W$ structure described in Section 5 where each of the variables is continuous. There are 14 network parameters: 5 conditional means, 5 conditional variances, and 4 linear parameters. The marginal distribution for the observed variables has 14 parameters, whereas the analysis of the rank of the Jacobian matrix tells us that the dimension of this model is 12. This coincides with the intuition that many values for the variance of $H$ and the linear parameters for $C \leftarrow H$ and $H \rightarrow D$ produce the same model for the observable variables, but once any two of these parameters are appropriately set, then the third parameter is uniquely determined by the marginal distribution for the observable variables.

## 8 Sigmoid Networks

Finally, let us consider the case where each of the variables $\{X_1, \ldots, X_n\} = \mathbf{X}$ is binary (discrete), and each local likelihood is the generalized linear model

$$p(x_i|\mathbf{pa}_i, \boldsymbol{\theta}_i, S) = \text{Sig}(a_i + \Sigma_{X_j \in \mathbf{Pa}_i} b_{ji} x_j)$$

where $\text{Sig}(x)$ is the *sigmoid function* $\text{Sig}(x) = \frac{1}{1+e^{-x}}$. These models, which we call *sigmoid networks*, are useful for learning relationships among discrete variables, because these models capture non-linear relationships among variables yet employ only a small number of parameters [Neal, 1992, Saul et al., 1996].

Using techniques similar to those in Section 5, we can compute the rank of the Jacobian matrix $\left[\frac{\partial \boldsymbol{\theta}_s}{\partial W}\right]$. We can not apply Thm. 1 to conclude that this rank is almost everywhere some fixed constant, because the local likelihoods are non-polynomial sigmoid functions. Nonetheless, the claim of Thm. 1 holds also for analytic transformations, hence a regular rank exists for sigmoid networks as well (as confirmed by our experiments).

Our experiments show expected reductions in rank for several sigmoid networks. For example, consider the two-level network

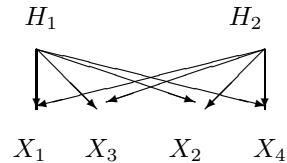

This network has 14 parameters. In each of 10 trials, we found the rank of the Jacobian matrix to be 14, indicating that this model has dimension 14. In contrast, consider the three-level network.

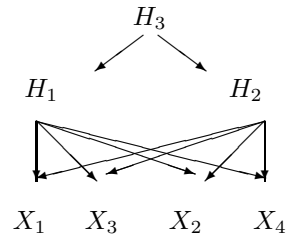

This network has 17 parameters, whereas the dimension we compute is 15. This reduction is expected, because we could encode the dependency between the two variables in the middle level by removing the variable in the top layer and adding an arc between these two variables, producing a network with 15 parameters.